# Context-sensitive Spelling Correction Using Google Web 1T 5-Gram Information

Youssef Bassil[1] & Mohammad Alwani[1]

[1] LACSC - Lebanese Association for Computational Sciences, Beirut, Lebanon

Correspondence: Youssef Bassil, LACSC - Lebanese Association for Computational Sciences, Registered under No. 957, 2011, Beirut, Lebanon. E-mail: youssef.bassil@lacsc.org

Received: January 9, 2012　　Accepted: February 11, 2012　　Online Published: May 1, 2012

doi:10.5539/cis.v5n3p37　　　　URL: http://dx.doi.org/10.5539/cis.v5n3p37

*The research is financed by the Lebanese Association for Computational Sciences (LACSC), No. 957, 2011, under the "Web-Scale Spell-Checking Research Project - WSSCRP2011"*

**Abstract**

In computing, spell checking is the process of detecting and sometimes providing spelling suggestions for incorrectly spelled words in a text. Basically, a spell checker is a computer program that uses a dictionary of words to perform spell checking. The bigger the dictionary is, the higher is the error detection rate. The fact that spell checkers are based on regular dictionaries, they suffer from data sparseness problem as they cannot capture large vocabulary of words including proper names, domain-specific terms, technical jargons, special acronyms, and terminologies. As a result, they exhibit low error detection rate and often fail to catch major errors in the text. This paper proposes a new context-sensitive spelling correction method for detecting and correcting non-word and real-word errors in digital text documents. The approach hinges around data statistics from Google Web 1T 5-gram data set which consists of a big volume of n-gram word sequences, extracted from the World Wide Web. Fundamentally, the proposed method comprises an error detector that detects misspellings, a candidate spellings generator based on a character 2-gram model that generates correction suggestions, and an error corrector that performs contextual error correction. Experiments conducted on a set of text documents from different domains and containing misspellings, showed an outstanding spelling error correction rate and a drastic reduction of both non-word and real-word errors. In a further study, the proposed algorithm is to be parallelized so as to lower the computational cost of the error detection and correction processes.

## 1. Introduction

For many years, computers have been solely exploited to solve mathematical and computational problems. Nevertheless, during the last few years, this trend has changed radically as the rapid booming of IT industry and advances in computing technologies gave birth to a new breed of computing applications. One of these applications is the processing of human languages, a field that is often known as natural language processing (NLP) or computational linguistics. Spell checking is a sub-field of computational linguistics whose function is to detect and sometimes correct words in a text that are not spelled correctly (Manning, Raghavan, & Schütze, 2008). In essence, a spell checker or spell corrector is a computer program often integrated with a word processor that performs spell checking and is majorly composed of a blend of three units: The error detector which flags misspelled words by validating them against a dictionary or a lexicon of words; The candidate spellings generator which provides alternative corrections for the detected errors; and the error corrector which selects the best candidate as a replacement for the detected error.

At heart, a spell checker/corrector is based on a built-in dictionary of words to detect errors, and on a corpus-based probabilistic model to perform error correction. However, with the dynamic growth of new words and terminologies entering the language, closed and conventional dictionaries are no more adequate to cover every single word in the vocabulary. Besides, traditional dictionaries rarely cover proper names, names of countries and regions, technical keywords, domain-specific terms, and acronyms. Consequently, there will be little data in these dictionaries to cover all words of the language, a problem known as data sparseness (Allison, Guthrie, & Guthrie, 2006). Basically, the richer vocabulary a dictionary or any source of text has, the more accurate, in general, is an NLP system (Banko & Brill, 2001). Likewise, a study conducted by the computational linguistics community showed that a generic colossal corpus such as the web, can most of the time overcome data sparseness problems (Kilgariff & Grefenstette, 2003).

This paper proposes a new spelling error detection and correction technique for computer text documents, based on data statistics from Google Web 1T 5-gram data set (Google Inc., 2006) which encompasses a massive volume of *n*-gram words ranging from unigrams (1-gram) mostly suitable for dictionary implementation and 5-gram word sequences emulating a universal text corpus, all extracted from the World Wide Web. Inherently, the proposed technique consists of several building blocks: An error detector that detects non-word errors using unigram statistics from Google Web 1T data set; a candidate spellings generator coupled with a character-based 2-gram model that generates candidate spellings for every detected error; and a context-sensitive error corrector that selects the best spelling candidate to replace the detected error using 5-gram statistics from Google Web 1T data set.

**2. State of the Art**

Practically, spelling errors in type written text vary between 1% and 3% (Grudin, 1983; Kukich, 1992) where 80% of them are usually caused by trivial editing operations such as insertion, deletion, substitution, and transposition (Damerau, 1964). Nonetheless, a different study (Peterson, 1986) pointed out that 94% of spelling errors are typically caused by such editing operations. In fact, spelling error detection and correction algorithms can be merely broken down into several types (Kukich, 1992): The non-word error detection which consists of detecting error words that are non-words, that is, words that cannot be found in a dictionary; The isolated-word error correction which consists of correcting non-word errors but looking at them in isolation, independently of their context; And the context-dependent or context-sensitive error detection and correction which consists of detecting and correcting errors according to their context in the sentence.

In fact, spelling error correction is not a new subject; It has been exploited by several researchers for over decades now. Several linguistic models and algorithms were proposed and experimented; The most prominent ones are the Noisy Channel model, the *n*-gram model, the edit distance algorithm, and the context-sensitive error correction algorithms.

*2.1 Noisy Channel Model*

The concept behind the noisy channel model is to consider a spelling error as a noisy signal that has been distorted somehow during transmission. The quintessence of this approach is that if one could know how the original word was distorted, it is then easy to find the actual correction (Jurafsky & Martin, 2008). The noisy channel model is a special case of Bayesian inference (Bayes, 1963) which is principally a classification-type model that inspects some observations and ranks them into appropriate classes and categories. Bledsoe and Browning (1959), and Mosteller and Wallace (1964) were the first among other researches to apply the Bayesian inference to detect misspellings in computer generated text.

Mathematically, the Bayesian model is a probabilistic model based on statistical assumptions and probability theory, namely the prior probability $P(w)$ and the likelihood probability $P(O|w)$ which are usually calculated by the following equation:

$$\hat{w} = \underset{w \in V}{\operatorname{argmax}} \frac{P(O|w)P(w)}{P(O)} = \underset{w \in V}{\operatorname{argmax}} P(O|w)P(w)$$

*P(w)* is called the prior probability and indicates the probability of *w* to occur in a specific corpus. *P(O|w)* is called the likelihood probability and denotes the probability of observing a misspelling *O* given that the correct word is *w*. *O* is the actual misspelled word and *w* is a potential spelling candidate. For every candidate, the product of *P(O|w)\*P(w)* is to be calculated; The candidate having the greatest product is to be selected as a correction for *O* and is denoted by *w'*.

The prior probability *P(w)* is straightforward as it is simply computed as $P(w) = C(w) + 0.5 / N + 0.5$, where *C(w)* is the frequency or the number of occurrence of the word *w* in the corpus, and *N* is the total number of words in the corpus. In order to avoid zero counts for *C(w)*, the value of 0.5 is added to the equation. On the other hand, the likelihood *P(O|w)* is harder to calculate than *P(w)* as it is imprecise to find the probability of a word to be misspelled, however, it can be estimated by calculating the probability of possible wrongful insertion, deletion, substitution, and transposition in general.

Experiments conducted by Kernighan, Church and Gale (1990) proved that the Bayesian model is not perfect as it can fail to correct spelling errors in some cases, for instance, falsely correcting the spelling error "acress" as "acres", while the original word is "actress".

*2.2 N-Gram Model*

The *n*-gram model has been so far applied in many linguistics problems such as spelling correction, speech recognition, and word sequence prediction. Principally, the *n*-gram is a probabilistic model originally proposed by Markov (1913) and later applied by Shannon (1948), Chomsky (1956), and Chomsky (1957) to predict the next word in a particular sequence of words. In short, an *n*-gram is simply a collocation of words that is *n* words long. For instance, "the cat" is a 2-gram sequence also referred to as bigram, "the cat is eating" is a 4-gram sequence, "the cat is eating food and drinking" is a 7-gram sequence, and so forth. Below are examples for 3-gram and 4-gram word sequences with their frequencies extracted from Google Web 1T *n*-gram data set (Google Inc., 2006).

3-gram word sequences:

- ceramics collectables collectibles (55)
- ceramics collectables fine (130)
- ceramics collected by (52)
- ceramics collectible pottery (50)
- ceramics collectibles cooking (45)

4-gramword sequences:

- serve as the incoming (92)
- serve as the incubator (99)
- serve as the independent (794)
- serve as the index (223)
- serve as the indication (72)
- serve as the indicator (120)

Unlike the prior probability *P(w)* which calculates the probability of a word *w* regardless of its surrounding words, the *n*-gram model calculates the conditional probability *P(w|s)* of a word *w* given the previous sequence of words *s*, that is, predicting the next word based on the preceding *n-1* words. For example, the conditional probability of *P(car|blue)* consists of calculating the probability of the whole sequence "blue car". Put differently, for the word "blue", the probability that the next word is "car" is to be computed.

Since it is too complicated to calculate the probability of a word given all previous sequence of words, the bigram or 2-gram model is rather used most of the time. It is denoted by $P(w_n|w_{n-1})$ denoting the probability of a word $w_n$ given the previous word $w_{n-1}$. For a sequence of bigrams, the probability is calculated as follows:

$$P(w_1^n) \approx \prod_{k=1}^{n} P(w_k | w_{k-1})$$

Several broader studies were investigated to improve the *n*-gram model from different perspectives; this may include but not limited to smoothing techniques suggested to solve the problem of zero-frequency of *n*-grams that never occurred in a corpus (Jeffreys, 1948; Church & Gale, 1991), the weighted *n*-gram model that more accurately estimates the *n*-grams based on their location in the context (Kuhn & Mori, 1990), and the variable length *n*-gram model (Niesler & Woodland, 1996) which varies the length of grams to attaint better overall system performance and compactness.

*2.3 Minimum Edit Distance*

The Minimum Edit Distance algorithm was first conceived by Wagner and Fischer (1974), and it is defined as the minimum number of edit operations needed to transform a string *x* into a string *y*. These operations are insertion, deletion, and substitution. In spelling correction, the purpose of the Minimum Edit Distance algorithm is to reduce the number of candidate spellings by eliminating the candidates with maximum edit distance as they are considered to share fewer characters with the spelling error than other candidates. There exist different edit distance algorithms: Levenshtein (Levenshtein, 1966), Hamming (Hamming, 1950) and the Longest Common Subsequence (Allison & Dix, 1986) algorithms.

The Levenshtein algorithm employs a weighting mechanism that assigns a cost of 1 to every performed edit operation irrespective of its type (insertion, deletion, or substitution). For example, the Levenshtein edit distance between "cat" and "dog" is 3 (substituting c by d, a by o, and t by g). The Levenshtein edit distance between "ping" and "rings" is 2 (substituting p by r, and inserting s at the end of ping).

The hamming distance is yet another algorithm for measuring the distance between two strings of the same length. It is calculated by finding the minimum number of substitutions required to transform string $x$ into string $y$. Practically, the Hamming distance between "ring" and "ping" is 1 (changing r to p), the hamming distance between "334223" and "331227" is 2 (changing 4 to 1 and 3 to 7), and the hamming distance between "ring" and "pings" is invalid because the strings are not of the same length.

Another popular technique for finding the distance between two words is the LCS short for Longest Common Subsequence. The idea pivots around finding the longest common subsequence of two strings. A subsequence is a series of characters, not necessarily consecutive, that appear from left to right in a string. Accordingly, the longest common subsequence of two strings is the maximum length of the mutual subsequence. For example, if $x$=AA*BDD*L*PS*T*TA*CF*M* and $y$=*BD*A*D*SAQ*PD*S*TA*ABC*M*E, then LCS is equal to *BDDPSTAM*.

*2.4 Context-sensitive Spelling Error Correction*

Context-sensitive spelling error correction is the task of detecting and correcting spelling errors that result in valid words, i.e. real-word errors. For instance, in the sentence "you should constantly backup your computer flies", the word "flies" is a real-word error mostly caused by a typographical mistake. Obviously, the writer didn't intend to mean that computer flies like planes, but he most probably meant "computer files". This slight confusion produced a real-word error that is actually valid in the English dictionary, however invalid with respect to the sentence in which it has occurred. Context-sensitive spelling error correction tries to detect and correct such real-word errors by inspecting their grammatical and semantic contexts. Error correction based on grammatical context or syntactic context, attempts to apply grammatical rules to detect misspellings, for instance, asserting that the word "play" in the sentence "he play" is a grammatical error is true since in the English language, a third person verb in the present tense must always ends with an "s". In contrast, error correction based on semantic context can correct the word "peace" into "piece" in the sentence "peace of cake". Since words "peace" and "piece" are valid nouns in the English language, they are hard to be flagged by traditional non context-sensitive spell checkers.

Mays, Damerau and Mercer (1991) proposed using the *n*-gram model to predict the actual correction for a real-word error. The idea centers around generating candidate spellings for every misspelled word by only applying simple edit operations such as insertion, deletion, and substitution, and then using *n*-gram statistics derived from a corpus to compute $P(w_n|w_{n-1})$.

Church and Gale (1991) suggested the use of a noisy channel to predict the actual correction for a real-word error. The technique harnesses a 100 million words corpus and *n*-gram statistics to correct errors according to their contextual information.

Liu and Curran (2006) employed *n*-gram statistics to correct real-word errors using a big corpus of text collected from crawling the web. As a result, huge improvements were achieved due to the large volume and generality of web corpuses.

Carlson and Fette (2007) employed the same previous technique but instead a memory-based learner was used to correct cross-domain errors. The system was trained using *n*-gram data tokens extracted from the web. The experiments yielded high precision real-word and non-word error correction.

Another approach was proposed by Demetriou, Atwell and Souter (1997), based on semantic knowledge and large vocabulary to correct spelling errors. A semantic model was built based on semantic association between words in a text to largely decrease the semantic ambiguities in natural languages.

Hodge and Austin (2003) proposed a supervised learning spell checking methodology based on Hamming distance algorithm and on an *n*-gram model for detecting isolated word errors. The generated candidate spellings are ranked based on their Hamming distance and *n*-gram statistics. In due course, candidates having the highest score are selected as correction for the detected real-word errors.

Golding and Roth (1999) applied a machine-learning approach and the Winnow algorithm, taking the surrounding words of the spelling error as features to resolve lexical disambiguation. The approach is based on features like part-of-speech to solve word disambiguation.

Islam and Inkpen (2009) presented a method for detecting real-word spelling errors based on data extracted from the web. The Google Web 1T *n*-gram data set was used to generate candidate spellings and a modified version of the Longest Common Subsequence (LCS) algorithm was used to select the best candidate to replace every single detected error.

**3. Proposed Solution**

This paper proposes a new context-sensitive spelling error correction method for detecting and correcting non-word and real-word errors in generic computer text documents. Fundamentally, the proposed method is a blend of three algorithms, each having a particular purpose and task. The task of the first algorithm is to detect non-word errors using Google Web 1T unigram data set (a subset of Google Web 1T 5-gram data set). The task of the second algorithm is to generate a list of candidate spellings for every detected error in the text using Google Web 1T unigram data set and a character-based 2-gram model. The task of the third algorithm is to perform context-sensitive error correction and select the best appropriate spelling candidate using 5-gram word counts from Google Web 1T 5-gram data set. Figure 1 depicts the logical block diagram for the three algorithms of the proposed solution.

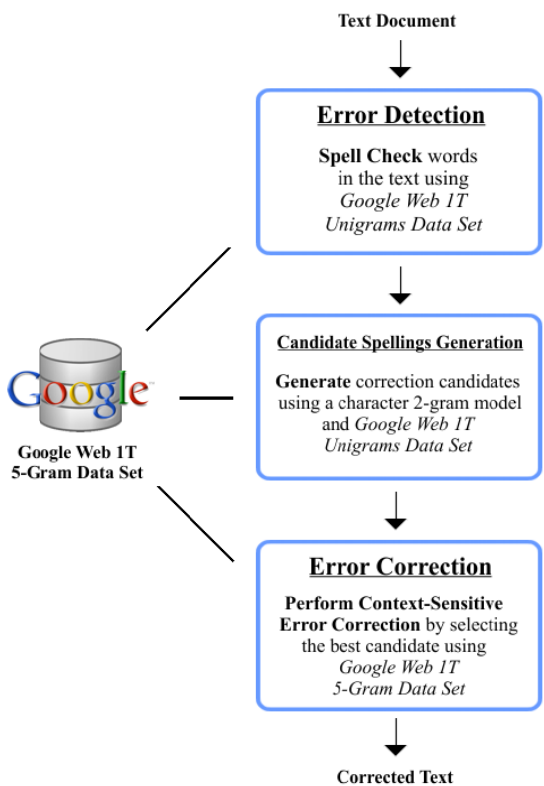

Figure 1. The proposed solution and its different components

*3.1 Google Web 1T 5-Gram Data Set*

Under the hood of a nice looking and feature-rich search engine such as Google, there actually exists a titanic database of millions of public web pages containing trillions of word collocations and word *n*-gram sequences, which can be used as a ground work for miscellaneous natural language processing applications such as machine translation, speech recognition, spell checking, as well as other types of computer-based linguistics problems. Google Inc. has already published at the Linguistic Data Consortium (LDC) its Web 1T 5-gram data set which is also available on six DVDs. This massive data set contains English word *n*-grams with their respective frequencies extracted from online public web pages. Table 1 delineates the number of *n*-gram words, up to 5-grams with their corresponding counts, published by Google Inc.

Table 1. Google Web 1T 5-gram data set

| File size: Approximately 24 GB Compressed Text Files | |
|---|---|
| Number of tokens | 1,024,908,267,229 |
| Number of sentences | 95,119,665,584 |
| Number of unigrams | 13,588,391 |
| Number of bigrams | 314,843,401 |
| Number of trigrams | 977,069,902 |
| Number of 4-grams | 1,313,818,354 |
| Number of 5-grams | 1,176,470,663 |

*3.2 Error Detection Algorithm*

The proposed error detection algorithm detects non-word errors $E=\{e_1,e_2,e_3,e_m\}$ in the original text $T=\{w_1,w_2,w_3,w_n\}$, where $e$ is a misspelled word or simply an error word, $m$ is the total number of detected errors, $w$ is a word in the original text and $n$ is the total number of words in the text. The process starts by validating every word $w_i$ in $T$ against Google's data set of unigrams; If $w_i$ is found, then $w_i$ is said to be correct and no correction is to take place. Otherwise, if the word $w_i$ is not found, then $w_i$ is said to be misspelled, and hence a correction is required. Google's data set of unigrams is already sorted alphabetically, and thus Binary search can be employed immaculately to speed up the execution time of error detection. Ultimately, a list of errors is generated and is denoted by $E=\{e_1, e_2, e_3, e_m\}$ where $m$ is the total number of non-word errors detected in the original text. Below is the pseudo-code for the proposed error detection algorithm.

```
Proc Error Detection (T)
{
    W ←  Split (T , " ")   // splits the text on every space and stores the returned words into an array W
    for(i←0 to N)    // iterates until all words are validated
    {
        // searches for every W[i] in Google Web 1T unigrams data set
        flag←BinarySearch (Google Unigrams Data Set, W[i])

        if(flag = true)    // indicates that W[i] is found in Google data set and thus it is spelled correctly
            i← i+1// move to the next word W[i+1]
        else    // indicates that W[i] is misspelled and thus it requires correction
            GenerateCandidates (W[i])// proceed with the candidate spellings generation algorithm
    }
}
```

*3.3 Candidate Spellings Generation Algorithm*

The proposed candidate spellings generation algorithm builds a list of possible spelling corrections for every detected non-word error. Those candidate corrections are denoted by $C=\{c_{11},c_{12},c_{13},c_{1r},...,c_{m1},c_{m2},c_{m3},c_{mq}\}$ where $c$ denotes a particular candidate spelling, $m$ denotes the total number of detected non-word errors, and $r$ and $q$ denote the total number of candidates generated for a particular detected error. In essence, the algorithm is based on a character-based 2-gram model which searches for unigrams in Google Web 1T data set having 2-gram character sequences in common with the error word.

For example, assuming that the original sentence to be validated is "case where only one sangle element is allowed to be stored" in which the word "single" was misspelled as "sangle", the non-word error "sangle" can be broken down into 2-gram character sequences as follows: sangle → sa, an, ng, gl, le.

Considering a sample list of unigrams from Google Web 1T data set such as: salute, sandbox, sand, sale, sandwich, salt, sanitary, tangle, man, angle, single, English, tingle, fringe, ring, singly, beagle, unable, disable.

Searching for unigrams in this list that share 2-gram character sequences with the error word "sangle", would give the following results:

sa: *sa*lute    *sa*ndbox    *sa*nd    *sa*le    *sa*ndwich    *sa*lt    *sa*nitary

an: t*an*gle    s*an*itary    s*an*dbox    s*an*d    s*an*dwich    m*anan*gle

ng: ta*ng*le    si*ng*le    E*ng*lish    a*ng*le    ti*ng*le    fri*ng*e    ri*ng*

gl: sin*gl*e    sin*gl*y    tin*gl*e    an*gl*e    bea*gl*e    tan*gl*e    En*gl*ish

le: sing*le*    ang*le*    beag*le*    unab*le*    ting*le*    tang*le*    disab*le*

The top 10 words having the highest number of common 2-gram character sequences with the error word "sangle" are selected as candidate spellings, and they are:

"tangle": It shares 4 sequences with "sangle"

"angle": It shares 4 sequences with "sangle"

"single": It shares 3 sequences with "sangle"

"tingle": It shares 3 sequences with "sangle"

"beagle": It shares 2 sequences with "sangle"

"sand": It shares 2 sequences with "sangle"

"sandbox": It shares 2 sequences with "sangle"

"English": It shares 2 sequences with "sangle"

"sanitary": It shares 2 sequences with "sangle"

"sandwich": It shares 2 sequences with "sangle"

Choosing the top 10 unigrams ensures that the correction word is most of the time in the candidates list. Unigrams having same number of common 2-gram character sequences are prioritized according to their length with the respect to the error word, for instance, "sangle" is made out of 6 characters; and hence, words whose length is 6 are favored over those whose length is 5 or 7. The complete pseudo-code for the proposed candidate spellings generation algorithm is given below:

```
Proc Generate Candidates (error)
{
        seq← Tokenize (error) // returns 2-gram character sequences and stores them in array seq

        for(i←0 to N) // iterates until all 2-gram character sequences in seq are processed
        {
                // searches for unigrams having seq[i] as substring i.e. unigrams sharing 2-gram character
                // sequences with the error word
                results[i] ← Substring (GoogleUnigramsDataSet, seq[i])

                        i←i+1
        }
        // selects the top 10 words in results having the highest number of common 2-gram character sequences
        // with the error word and stores them into array candidates
        candidates←MAX (GetCommonWords (results))

        Error Correction (candidates) // launches the error correction algorithm
}
```

Based on the above results, the generated candidate spellings are $C_{sangle}$={tangle, angle, single, tingle, beagle, sand, sandbox, English, sanitary, sandwich}. Now, the ultimate task is to select the best candidate to replace the error word "sangle", a task left for the third algorithm namely the context-sensitive real-word error correction.

*3.4 The Context-sensitive Error Correction Algorithm*

The proposed context-sensitive spelling error correction algorithm takes each generated candidate $c_{ik}$ with four of the words that precede the original error in the original text, leading to $S_k$="$w_{i-4}w_{i-3}w_{i-2}w_{i-1}\ c_{ik}$" where S denotes a 5-gram word sentence, *w* denotes a word preceding the original error, *c* denotes a particular candidate spelling for a particular error, *i* denotes the $i_{th}$ word preceding the original error, and *k* denotes the $k_{th}$ candidate spelling. Each constructed sentence $S_k$ is then compared with Google 5-gram word counts from Google Web 1T 5-gram data set. The candidate $c_{ik}$ that belongs to the sentence $S_k$ with the highest count is selected as a replacement for originally detected error word.

Back to the previous example, the list of S sentencescan be outlined as follows:

$S_1$= "case where only one tangle"

$S_2$= "case where only one angle"

$S_3$= "case where only one single"

$S_4$= "case where only one tingle"

$S_5$= "case where only one beagle"

$S_6$= "case where only one sand"

$S_7$= "case where only one sandbox"

$S_8$= "case where only one English"

$S_9$= "case where only one sanitary"

$S_{10}$= "case where only one sandwich"

The candidate spelling $c_k$ (tangle, angle, single, etc.) in the sentence $S_k$ having the highest frequency in Google Web 1T 5-gram data set is selected as a correction for the error word "sangle". The proposed algorithm is context-sensitive as it relies on real world word counts from Google data set initially extracted from the Internet. Therefore, despite the fact that the word "single" might be misspelled as the real-word "tingle", the algorithm should be able to correct it since the sentence "case where only one tingle" is to occur very few times over the Internet, fewer than any other sentence, for instance, "case where only one single".

Next is the pseudo-code for the proposed context-sensitive error correction algorithm.

```
Proc Error Correction (candidates)
{
    for(i←0 to N)    // iterates until all words are validated
    {
        // concatenates the ith candidate with the four words that precede the error
        // W and k are supposed to be global variables
        S ←Concat (W[k-4], W[k-3], W[k-2], W[k-1], candidates[i])

        // searches for S in Google 5-gram data set and returns its frequency
        count[i] ←BinarySearch (Google 5 grams Data Set, S)
        i ← i+1
    }

    index←MAX (count)    // returns the index of the candidate whose S has the highest frequency

    RETURN candidates [index]    // returns the correction for the misspelled word
}
```

## 4. Experiments and Results

For evaluation purposes, 300 articles pertaining to various domains were experimented including finance, business, IT, literature, political science, medicine, sports, and others. In total, they comprise 200,000 words including dictionary words, proper names, domain specific terms and terminologies, acronyms, and technical jargons and expressions. Initially, those articles are error-free as they do not contain any misspellings or linguistic mistakes; however, several words were randomly altered on purpose yielding to non-word and real-word errors in the text. These induced errors are approximately 1% of the original text; and hence they are around 2,000 spelling errors. Table 2 gives the total number of words in the set of articles, in addition to the number of induced non-word and real-word errors.

Table 2. Number of induced errors

| Total Words | Total Errors    | Non-Word Errors | Real-Word Errors |
|-------------|-----------------|-----------------|------------------|
| 200,000     | 2,000           | 1,600           | 400              |
|             | 1% of 200,000   | 80% of 2,000    | 20% of 2,000     |

For comparison purposes, the GNU Aspell (Atkinson, 2004) and Ghotit Dyslexia (Ghotit ltd., 2011) were used to spell check the test data. The Aspell is a free-software cross-platform spell checker that is the standard spell checker for the GNU software project and has been integrated into commercial software applications such as Notepad++, Opera, gedit, and others. It is compatible with Unix-based operating systems, as well as Microsoft Windows. On the other hand, Ghotit Dyslexia is a proprietary contextual spell checker developed by Ghotit and mostly intended for people with dyslexia, dysgraphia, and other English writing difficulties. Ghotit is a Microsoft Word add-on that includes a context spell checker, a grammar checker, and an integrated word dictionary.

The results of executing Aspell to spell check the test data are given in Table 3, while the results for Ghotit are given in Table 4.

Table 3. GNU Aspell test results

| Total Errors | | Non-Word Errors | | Real-Word Errors | |
|---|---|---|---|---|---|
| 2,000 | | 1,600 | | 400 | |
| 1% of 200,000 total words | | 80% of 2,000 | | 20% of 2,000 | |
| Corrected | Not/Falsely Corrected | Corrected | Not/Falsely Corrected | Corrected | Not/Falsely Corrected |
| 1,020 | 980 | 988 | 612 | 32 | 368 |
| 51% of 2,000 | 49% of 2,000 | 62% of 1,600 | 38% of 1,600 | 8% of 400 | 92% of 400 |

Table 4. Ghotit test results

| Total Errors | | Non-Word Errors | | Real-Word Errors | |
|---|---|---|---|---|---|
| 2,000 | | 1,600 | | 400 | |
| 1% of 200,000 total words | | 80% of 2,000 | | 20% of 2,000 | |
| Corrected | Not/Falsely Corrected | Corrected | Not/Falsely Corrected | Corrected | Not/Falsely Corrected |
| 1,234 | 766 | 1,118 | 482 | 116 | 284 |
| 62% of 2,000 | 38% of 2,000 | 70% of 1,600 | 30% of 1,600 | 29% of 400 | 71% of 400 |

Applying the proposed method on the test data to detect and correct spelling errors resulted in 1,860 errors being corrected successfully, among which 1,581 were non-word errors and 279 were real-word errors. As a result, around 93% of the total errors were corrected; around 99% of total non-word errors were corrected; and around 70% of total real-word errors were corrected successfully. Table 5 outlines the obtained test results for the proposed method.

Table 5. Proposed method test results

| Total Errors | | Non-Word Errors | | Real-Word Errors | |
|---|---|---|---|---|---|
| 2,000 | | 1,600 | | 400 | |
| 1% of 200,000 total words | | 80% of 2,000 | | 20% of 2,000 | |
| Corrected | Not/Falsely Corrected | Corrected | Not/Falsely Corrected | Corrected | Not/Falsely Corrected |
| 1,860 | 140 | 1,581 | 19 | 279 | 121 |
| 93% of 2,000 | 7% of 2,000 | 99% of 1,600 | 1% of 1,600 | 70% of 400 | 30% of 400 |

Below are examples of successful and unsuccessful corrections observed during the execution of the proposed error correction method. It is worth noting that errors are marked by an underline and results are interpreted using a special notation in the form of [error-type ; corrected error ; intended word].

Successfully Corrected 93%:

… would like to ask you to voice your sopport for this bill …→ [non-word error; support; support]

… but the content of a computer is vulnerable to fee risks …→ [real-word error; few; few]

… medical errors effect us all whether we are involved or not …→ [real-word error; affect; affect]

… many of the best poems are found in too collections …→ [real-word error; two; two]

Not Corrected 2%:

… whether you hit the road in a sleek imported sporting car …→ [real-word error; sporting; sports]

… we fear the precaution of medication prior to tonsillectomy …→ [real-word error; fear; feel]

Falsely Corrected 4%:

… After all I slept near my door on the pavement…→ [real-word error; dog; door]

… I saw the ball running too fast …→ [real-word error; bus; ball]

In a head-to-head comparison, it is evident that the proposed method outperformed the other two existing solutions as a higher number of non-word and real-word errors were detected and corrected successfully. Particularly, the proposed method managed to correct 99% of total non-word errors and 70% of total real-word errors, yielding to an error correction rate close to 93%. Only 7% of total errors were left either undetected or were falsely corrected. In contrast, the GNU Aspell yielded an error correction rate of 51%, while the Ghotit yielded a 62% rate. It was obvious that the strong point of the proposed method was in real-word error correction (context-sensitive) as it outscored the Aspell spell checker by 800% (8 times more errors were corrected), and the Ghotit spell checker by 240% (2.4 times more errors were corrected). These outstanding results are primarily due to the large count of 5-gram tokens and their abundant statistics in the Google Web 1T data set harnessed by the proposed method. Moreover, since the data of Google Web 1T set are pulled out of the Internet, it is heavily stuffed with real data encompassing dictionary words, proper names, domain specific terms and terminologies, acronyms, and technical jargons and expressions that can cover most of the words and their possible sequences in the language.

**5. Conclusions and Future Work**

This paper presented a novel context-sensitive approach for detecting and correcting non-word and real-word spelling errors in text documents. The proposed algorithm is based on Google Web 1T 5-gram data set that houses a huge volume of word sequences originally extracted from Internet web pages. The goal of this new method was to improve the error correction rate of modern spell checkers, especially context-sensitive error correction. The proposed method excelled when put under test alongside with other spell checkers, more particularly the GNU Aspell and the proprietary Ghotit Dyslexia. In effect, 99% non-word errors and 70% real-word errors were corrected by the proposed method; While the closest competitor namely Ghotit hit approximately 70% for non-word errors and 29% for real-word errors. Overall, 93% of total errors were corrected by the proposed method, while Ghotit scored 62%. In a nutshell, the proposed method was able to detect and correct 2.4 times more errors than the best existing method. The major reason behind these noteworthy results is the integration of Google Web 1T data set into the proposed algorithm as it embraces a wide-ranging set of words and precise statistics about word associations that cover domain-specific terms, technical terminologies, acronyms, expressions, proper names, and almost every word in the language.

As for future work, a parallel algorithm is to be devised and experimented; It can typically be implemented over multiprocessor machines or distributed computing infrastructures with the purpose of boosting the execution time and performance of the error detection and correction processes.

**References**

Allison, L., & Dix, T. I. (1986). A bit-string longest common-subsequence algorithm. *Information Processing Letters, 23,* 305-310. http://dx.doi.org/10.1016/0020-0190(86)90091-8

Allison, B., Guthrie, D., & Guthrie, L. (2006). Another Look at the Data Sparsity Problem. *Proceedings of the 9th International Conference on Text, Speech and Dialogue*, Czech Republic. http://dx.doi.org/10.1007/11846406_41

Atkinson, K. (2004). *GNU Aspell Spell Checker*. Retrieved from http://aspell.net/

Banko, M., & Brill, E. (2001). Scaling to Very Very Large Corpora for Natural Language Disambiguation. *In ACL*, pp. 26-33. http://dx.doi.org/10.3115/1073012.1073017

Bayes, T. (1963). An Essay Toward Solving a Problem in the Doctrine of Chances. reprinted in *Facsimiles of two papers by Bayes, Hafner Publishing Company, 53,* New York.

Bledsoe, W. W., & Browning, I. (1959). Pattern recognition and reading by machine. *Proceedings of the Eastern Joint Computer Conference*, pp. 225-232, Academic, New York. http://dx.doi.org/10.1145/1460299.1460326

Carlson, A., & Fette, I. (2007). Memory-Based Context-Sensitive Spelling Correction at Web Scale. *Proceedings of the IEEE International Conference on Machine Learning and Applications (ICMLA)*.

Chomsky, N. (1956). Three models for the description of language. *IRI Transactions on Information Theory*, *2*(3), 113-124. http://dx.doi.org/10.1109/TIT.1956.1056813

Chomsky, N. (1957). *Syntactic Structures.* Mouton, The Hague.

Church, K. W., & Gale, W. A. (1991). A comparison of the enhanced Good-Turing and deleted estimation methods for estimating probabilities of English bigrams. *computer speech and language, 5,* 19-54. http://dx.doi.org/10.1016/0885-2308(91)90016-J

Church, K. W., & Gale, W. A. (1991). Probability scoring for spelling correction. *Statistics and Computing*. http://dx.doi.org/10.1007/BF01889984

Damerau, F. J. (1964). A technique for computer detection and correction of spelling errors. *Communications of the ACM*, *7*(3), 171-176. http://dx.doi.org/10.1145/363958.363994

Demetriou, G., Atwell, E., & Souter, C. (1997). Large-scale lexical semantics for speech recognition support. *In EUROSPEECH*, pp. 2755-2758.

Ghotit ltd. (2011). *Ghotit Dyslexia Contextual Spell Checker.* Retrieved from http://www.ghotit.com/home.shtml

Golding, A. R., & Roth, D. (1999). A winnow-based approach to context-sensitive spelling correction. *Machine Learning*, *34*(3), 107-130. http://dx.doi.org/10.1023/A:1007545901558

Google Inc. (2006). *Google Web 1T 5-Gram Version 1*. Retrieved from http://www.ldc.upenn.edu/Catalog/CatalogEntry.jsp?catalogId=LDC2006T13

Grudin, J. T. (1983). Error patterns in novice and skilled transcription typing. Cooper edition, Springer-Verlag, *Cognitive Aspects of Skilled Typewriting*, pp. 121-139.

Hamming, R. W. (1950). Error detecting and error correcting codes. *Bell System Technical Journal, 29*(2), 147-160.

Hodge, V. J., & Austin, J. (2003). A comparison of standard spell checking algorithms and novel binary neural approach. *IEEE Trans. Know. Dat. Eng.*, *15*(5), 1073-1081. http://dx.doi.org/10.1109/TKDE.2003.1232265

Islam, A., & Inkpen, D. (2009). Real-Word Spelling Correction using Google Web 1T n-gram Data Set. *in Proceedings of the 18th ACM Conference on Information and Knowledge Management*, Hong Kong, pp. 1689-1692. http://dx.doi.org/10.1145/1645953.1646205

Jeffreys, H. (1948). *Theory of Probability* (2nd ed.). Clarendon Press, Oxford.

Jurafsky, D., & Martin, J. (2008). *Speech and Language Processing* (2nd ed.). Prentice Hall.

Kernighan, M. D., Church, K. W., & Gale, W. A., (1990). A spelling correction program base on a noisy channel model. *In COLING-90*, Helsinki, 2(1), 205-211. http://dx.doi.org/10.3115/997939.997975

Kilgariff, A., & Grefenstette, G. (2003). Introduction to the Special Issue on the Web as a Corpus. *journal of computational linguistics*, MIT Press, p. 29. http://dx.doi.org/10.1162/089120103322711569

Kuhn, R., & De, M. R. (1990). A cache-based natural language model for speech recognition. *IEEE Transactions on Pattern Analysis and Machine Intelligence*, 12(6), 570-583. http://dx.doi.org/10.1109/34.56193

Kukich, K. (1992). Techniques for automatically correcting words in text. *ACM Computing Surveys*, 24(4), 377-439. http://dx.doi.org/10.1145/146370.146380

Levenshtein, V. I. (1966). Binary codes capable of correcting deletions, insertions, and reversals. *Cybernetics and Control Theory*, 10(8), 707-710.

Liu, V., & Curran, J. R. (2006). Web text corpus for natural language processing. *In EACL, The Association for Computer Linguistics*.

Raghavan, M., & Schütze (2008). *An Introduction to Information Retrieval*. Cambridge University Press.

Markov, A. A. (1913). Essaid'unerecherchestatistiquesur le texte du roman "EugèneOneguine", *Bull. Acad. Imper. Sci.,* St. Petersburg. http://dx.doi.org/10.1017/S0269889706001074

Mays, E., Damerau, F. J., & Mercer, R. L. (1991). Context based spelling correction. *Information Processing and Management*, 27(5), 517-522. http://dx.doi.org/10.1016/0306-4573(91)90066-U

Mosteller, F., & Wallace, D. L. (1964). *Inference and disputed authorship*: The Federalist. Springer-Verlag, New York. http://dx.doi.org/10.1214/aoms/1177699628

Niesler, T. R., & Woodland, P. C. (1996). A variable-length Category-based n-gram language model. *In IEEE ICASSP-96*, Atlanta, GA, 1, pp. 164-167. http://dx.doi.org/10.1109/ICASSP.1996.540316

Peterson, J. L. (1986). A note on undetected typing errors. *Communications of the ACM*, 29(7), 633-637. http://dx.doi.org/10.1145/6138.6146

Shannon, C. E. (1948). A mathematical theory of communication. *Bell system Technical Journal*, 27(3), 379-423.

Wagner, R. A., & Fischer, M. J. (1974). The string-to-string correction problem. *Journal of the Association for Computing Machinery*, 21, 168-173. http://dx.doi.org/10.1145/321796.321811